\documentclass{article}

\usepackage{amsmath}
\usepackage{amsfonts}
\usepackage{amssymb}
\usepackage{color}  
\usepackage{amsthm}
\usepackage{graphicx}
\usepackage{caption}
\usepackage{comment}
\usepackage{subcaption}
\usepackage{algpseudocode}
\usepackage{algorithm}
\usepackage[utf8]{inputenc}
\usepackage{tabularx}
\usepackage{multirow}
\usepackage{hyperref}

\usepackage{natbib}
\bibpunct[, ]{(}{)}{,}{a}{}{,}%

\title{Relaxed Dual Optimal Inequalities for Relaxed Columns:  with Application to Vehicle Routing}

\author{Naveed Haghani\textsuperscript{\rm 1}, Claudio Contardo\textsuperscript{\rm 2},  Julian Yarkony\textsuperscript{\rm 3}\\[2ex] 
\textsuperscript{\rm 1}University of Maryland, College Park, MD\\ 
\textsuperscript{\rm 2}ESG UQAM and GERAD, Montreal, Canada\\ 
\textsuperscript{\rm 3}Verisk Computational and Human Intelligence Laboratory, Jersey City, NJ\\ 
}\date{April 2019}

\begin{document}

\maketitle
\begin{abstract}
 We address the problem of accelerating column generation for set cover problems in which we relax the state space of the columns to do efficient pricing.  We achieve this by adapting the recently introduced smooth and flexible dual optimal inequalities (DOI) for use with relaxed columns.  Smooth DOI exploit the observation that similar items are nearly fungible, and hence should be associated with similarly valued dual variables.  Flexible DOI exploit the observation that the change in cost of a column induced by removing an item can be bounded.  We adapt these DOI to the problem of capacitated vehicle routing in the context of ng-route relaxations.  We demonstrate significant speed ups on a benchmark data set, while provably not weakening the relaxation.
\end{abstract}
\section{Introduction}
In this paper we accelerate the column generation (CG) solution to expanded linear programming (LP) relaxations \citep{barnprice} using dual optimal inequalities \citep{ben2006dual} (DOI). 
Expanded LP relaxations are used to solve integer linear programs (ILPs) for which compact LP relaxations are very loose.  In contrast to compact LP relaxations, which contain a small number of variables,  expanded LP relaxations contain a massive number of variables (called columns).  However an expanded LP relaxation is often much tighter than the corresponding compact LP relaxation, and permits efficient (often in practice exact) optimization \citep{mwspJournal} of the corresponding ILP.  To solve expanded LP relaxations, CG is used.  Since the set of all feasible columns is enormous and can not be easily enumerated, a sufficient set is constructed iteratively using CG.  The process of identifying negative reduced cost columns is called pricing.   Pricing operates by  via solving a small scale combinatorial optimization problem parameterized by the dual solution of the expanded LP relaxation defined over the nascent set of columns. 

CG has classically been accelerated using application specific DOI \citep{ben2006dual}, which operate by decreasing the size of the space of dual solutions that CG searches over.  DOI are constraints on the space of dual solutions that do not change the objective of the optimal primal/dual solution produced at termination of CG.  
%
Recently more general classes of DOI have been established, that accelerate CG optimization for applications in computer vision \citep{HPlanarCC,mwspJournal,wang2018accelerating}, entity resolution \citep{FlexDOIArticle} and operations research \citep{haghani2020smooth}.  In \citep{haghani2020smooth} the flexible DOI (F-DOI) framework of \citep{FlexDOIArticle} is adapted to describe rebates for over-including customers.   \citep{haghani2020smooth} observes that similar customers (e.g. with regards to spatial position and demand) should be associated with similar dual values resulting in Smooth DOI (S-DOI).  In \citep{haghani2020smooth} the combination of S-DOI and F-DOI is referred to as SF-DOI, and is tested on single source capacitated facility location \citep{holmberg1999exact}.  SF-DOI provide up to 130 times speed up for the problems considered in \citep{haghani2020smooth}, while provably not changing the final solution. 

Some classes of mixed-integer optimization problems such as the capacitated vehicle routing problem (CVRP) consider a super-set of the columns to increase tractability of pricing.  We refer to the additional columns/original columns as relaxed columns and valid columns respectively. One key problem with SF-DOI is that they do not model relaxed columns.  For CVRP a column is valid if  (1) it includes no customer more than once (no cycles in the corresponding route), and (2) does not service more demand than the vehicle has capacity.  For CVRP the pricing problem is an elementary resource constrained shortest path problem, which is strongly NP-hard \citep{desrosiers2005primer}.  To make solving large CVRP instances feasible, a super-set of the set of valid columns is often considered called ng-routes \citep{baldacci2011new}.  The use of ng-routes makes pricing tractable at the cost of some decrease in the tightness of the underlying expanded LP relaxation.  The ng-route relaxation permits a customer to be visited more than once in a route but forbids many short cycles localized in space. 

In this paper we make the following contributions to the scientific literature.  (1) We provide valid DOI for CVRP via the SF-DOI framework. (2) We show that the SF-DOI accelerate CG optimization over the ng-route relaxed columns.

We organize this document as follows. In Section \ref{sec_prelim} we provide background on CG, CVRP, DOI and ng-routes needed to understand the remainder of the document.  In Section \ref{Sec_DOI_valid_CVRP} we provide SF-DOI for CVRP problems.  In Section \ref{sec_exper} we consider the effectiveness of our SF-DOI on benchmark CVRP problems.  We conclude in Section \ref{sec_conc}.  
\section{Preliminaries}
\label{sec_prelim}
In this section we review the minimum weight set cover formulation used in operations research along with the CG solution and its application to vehicle routing \citep{Desrochers1992}.  We organize this section as follows.  In Section \ref{coverForm} we introduce minimum weight set cover for general problems and consider the CG solution.  In Section \ref{doiReview} we consider the use of SF-DOI to accelerate CG \citep{haghani2020smooth}.  In Section \ref{invalidCOl} we consider the use of relaxed columns, which are used when pricing in CG is intractable.  In Section \ref{invalid_doi} we consider the use of approximations to DOI in the case where DOI can not be computed.  In Section \ref{vehicleROuting} we  formally describe CVRP.  In Section \ref{ng_route_overview} we consider the use of relaxed columns for CVRP in the form of ng-routes.  
\subsection{Optimization for Minimum Weight Set Cover}
\label{coverForm}
We now consider the standard CG method for a minimum weight set cover problem. We are given a set of $N \in \mathbb{Z}_{+}$ items to be covered, denoted $\mathcal{N}$. The CG formulation includes a continuous variable $\theta_l\geq 0$ for every column $l\in\Omega$ where $\Omega$ is the set of all valid columns. In the standard cover formulation a column can cover an item at most once. This will be relaxed when we consider relaxed columns such as ng-routes \citep{baldacci2011new}, which may cover an item more than once.  In the context of CG, the set $\Omega$ typically is exponentially large with respect to $N$, and therefore it is impractical to explicitly consider the entire set during optimization. For every column $l\in\Omega$ and for every item $u\in \mathcal{N}$ we let $a_{ul}\in\{0, 1\}$ be a binary constant equal to 1 if $l$ covers $u$, and otherwise let $a_{ul}=0$. We associate a cost $c_l$ to the column $l$, via a non-decreasing function over $a_{ul}$ $\forall u \in \mathcal{N}$.  We write minimum weight set cover as:
\begin{align}
\label{setcover:cost}
    \min_{\theta \geq 0}\qquad \qquad \sum_{l\in\Omega} c_l\theta_l\\
 \sum_{l\in\Omega}a_{ul}\theta_l \geq 1 \quad \forall u\in \mathcal{N}
 \nonumber 
\end{align}
Given the unscalability of enumerating the set $\Omega$ explicitly, CG considers a subset $\Omega_R\subset\Omega$ at any given time, and thus the optimization (referred to as the restricted master problem or RMP) obtains the solution of the problem in Eq \ref{setcover:cost} but restricted to the columns in $\Omega_R$. Let us denote that problem $\mathtt{RMP}(\Omega_R)$. Now let us denote $(\alpha_u)_{u\in \mathcal{N}}$ as the dual variables associated with the constraints $\sum_{l\in\Omega}a_{ul}\theta_l \geq 1$ in problem $\mathtt{RMP}(\Omega_R)$. The reduced cost of a column $l\in\Omega\setminus\Omega_R$ can be computed as $\overline{c_l} = c_l - \sum_{u\in \mathcal{N}}\alpha_u a_{ul}$. Problem $\mathtt{RMP}(\Omega_R)$ provides a proven optimal solution of Eq \ref{setcover:cost} if $\min\{\overline{c_l} : l\in\Omega\setminus\Omega_R\} \geq 0$. Finding negative reduced cost columns is application specific but is generally a small scale combinatorial optimization problem.
\subsection{Dual Optimal Inequalities }
\label{doiReview}
Dual optimal inequalities (DOI) provide bounds on the dual variables, which provably do not remove all dual optimal solutions.  DOI are used to decrease the search space over $\alpha$, and hence dramatically accelerate optimization \citep{ben2006dual}.  
DOI have been historically only applied to specific problems with specially tailored structure \citep{ben2006dual} such as the cutting stock problem.  However recently \citep{mwspJournal,FlexDOIArticle,haghani2020smooth} showed that DOI can be constructed for very general classes of problems.  We now consider the smooth and flexible DOI (SF-DOI) of \citep{haghani2020smooth} in the context of minimum weight set cover formulations.
\subsubsection{Smooth-DOI}
S-DOI formalize the intuition that the similar items are nearly fungible, and hence their dual variables should have similar values. We now mathematically describe the S-DOI.  Let $\Omega_u$ be the set of columns in $\Omega$ including item $u$.  For any given $l \in \Omega,u\in \mathcal{N},v \in \mathcal{N}$ s.t. $l \in \Omega_u-\Omega_v$ let $\hat{l}=$swap$(l,u,v)$ be the column corresponding to replacing $u$ with $v$ in $l$.  Further let $\rho_{uv}$ be an upper bound on the amount that any column including $u$ but not $v$ increases in cost when $u$ is replaced by $v$.  Formally $\rho_{uv}$ satisfies:
\begin{align}
\label{smoothDOiDef}
\rho_{uv} \geq c_{\hat{l}}-c_{l} \quad \forall  l \in \Omega_u-\Omega_v, \hat{l}=\mbox{swap}(l,u,v)  
\end{align}
Given $\rho_{uv}$ defined in Eq \ref{smoothDOiDef} it is established in \citep{haghani2020smooth} that the dual values $\alpha$ can be bounded as follows without weakening the relaxation in Eq \ref{setcover:cost}.
\begin{align}
\label{rho_uv_def}
    \rho_{uv}\geq \alpha_v-\alpha_u \quad \forall u \in \mathcal{N}, v \in \mathcal{N}, u \neq v
\end{align}
Note that if the swap operation makes a column invalid (not lie in $\Omega$) then the cost of the resultant column is regarded as infinite.    We use  $\mathcal{S}$ to denote the set of pairs $u,v$ where $\rho_{uv}<\infty$.  We use $\mathcal{S}^-_u$ to denote the subset of $\mathcal{S}$ consisting of $(u,v)$ for all $v \in \mathcal{N}$.  Similarly we use $\mathcal{S}^+_u$ to denote the subset of $\mathcal{S}$ consisting of $(v,u)$ for all $v \in \mathcal{N}$.  

In the primal LP including S-DOI, which we detail in Section \ref{smooth_flex_doi_sec}, S-DOI permit items to be uncovered in exchange for a penalty being paid and other items being over-covered.  The primal LP introduces variables of the form  $\omega_s$ ($\forall s \in \mathcal{S}$) for $s=(u,v)$, which can be understood as counting the number of times $u$ is swapped for $v$ at cost $\rho_{uv}$.
\subsubsection{Flexible-DOI}
\label{flex_doi_desc}
F-DOI exploit the observation that if any item $u$ is included more than once in a solution to the RMP then that primal solution can be altered to remove excess of item $u$ from columns while decreasing the objective and preserving feasibility.  We describe this formally as follows.  For a column $l\in \Omega$,$u \in \mathcal{N}$ we define $\sigma_{ul}$ to satisfy the following properties.  \textbf{(1).}  $\sigma_{ul}\geq 0$.  \textbf{(2)}. $a_{ul} = 0\Rightarrow \sigma_{ul} = 0$. \textbf{(3)} The equation below is satisfied, which uses remove$(l, \hat{\mathcal{N}})$ to denote the column constructed by removing $\hat{\mathcal{N}}$ from $l$; where $\hat{\mathcal{N}}$ is a subset of $\mathcal{N}_l$ which is the set of items composing $l$.   
\begin{align}
    \label{criter3}
    \sum_{u\in \hat{\mathcal{N}}}\sigma_{ul}\leq c_l - c_{l'} 
    \quad \quad \forall l \in \Omega, \hat{\mathcal{N}}\subseteq \mathcal{N}_l = \{u\in \mathcal{N}: a_{ul} = 1\} \quad, l' =\mbox{remove}(l, \hat{\mathcal{N}}) 
\end{align}

Below we provide a sufficient condition to satisfy the requirement in Eq \ref{criter3} (see Appendix \ref{appendProof} for proof). 
\begin{align}
\label{altBoundFDOI}
    \sigma_{ul}\leq \min_{\substack{\mathcal{N}_{\hat{l}} \subseteq \mathcal{N}_l\\ \hat{l} \in \Omega_u\\ \bar{l}=\mbox{remove}(\hat{l},\{u\})}}
    c_{\hat{l}}-c_{\bar{l}}\quad \quad \forall u \in \mathcal{N}, l \in \Omega_u, 
\end{align}
Using $\sigma$ the primal RMP is augmented with additional variables, which permit the removal of items from columns in exchange for rebates provided according to $\sigma$.
The rebate for over-covering item $u$ using column $l$ is $\sigma_{ul}$.  
The primal LP introduces variables of the form  $\xi_{u\sigma}$, which counts the number of times $u$ is removed from a column for rebate $\sigma$.  This primal form is detailed in Section \ref{smooth_flex_doi_sec}. 

\subsubsection{Smooth and Flexible DOI in Optimization} 
\label{smooth_flex_doi_sec}
In addition to the notation introduced in Section \ref{coverForm}, let us introduce non-negative variables $\xi_{u\sigma}$ for every $u\in \mathcal{N}$ and for every $\sigma\in\Lambda_u$, where $\Lambda_u = \{\sigma_{ul}: l\in\Omega_R\}$ is the set of all possible values of $\sigma_{ul}$ across $\Omega_R$. The variable $\xi_{u\sigma}$ represents the number of columns $l$ from which $u$ will be removed with a rebate of $\sigma$. We let $\beta_{ul\sigma}$ be a binary constant equal to 1 if $\sigma_{ul} = \sigma$, and otherwise $\beta_{ul\sigma}=0$.    

We add one non-negative variable $\omega_s$ for every $s\in\mathcal{S}$.  Here $\omega_s$ denotes the number of times swap $s$ is applied.  We now formulate minimum weight set cover augmented with SF-DOI as optimization.  
\begin{align}\label{sfdoi:cost}
\min_{\substack{\theta \geq 0\\ \omega \geq 0\\ \xi \geq 0}}\qquad  \sum_{l\in\Omega} c_l\theta_l + \sum_{s\in\mathcal{S}}\rho_s\omega_s - \sum_{\substack{u\in \mathcal{N} \\ \sigma\in\Lambda_u}}\sigma\xi_{u\sigma}\\
\mbox{s.t.} \quad 
\xi_{u\sigma} - \sum_{l\in\Omega}\beta_{ul\sigma}\theta_l \leq 0  \quad \forall u\in \mathcal{N}, \sigma\in\Lambda_u \nonumber\\
 \sum_{l\in\Omega}a_{ul}\theta_l + \sum_{s\in \mathcal{S}^+_u} \omega_s - \sum_{s\in \mathcal{S}^-_u} \omega_s - \sum_{\sigma\in\Lambda_u} \xi_{u\sigma} \geq 1 \quad \forall  u\in \mathcal{N} \nonumber
\end{align}
It is established in \citep{haghani2020smooth} that the SF-DOI do not weaken the LP relaxation meaning that  Eq \ref{setcover:cost}= Eq \ref{sfdoi:cost}.  
\subsubsection{Efficient 
Implementation}
In this section we decrease the number of primal variables and primal constraints in the form in Eq \ref{sfdoi:cost} 
thus accelerating optimization without loosening the LP relaxation.  

\textbf{Accelerating F-DOI:  }Optimization over $\Omega$ may contain a massive number of $\xi$ variables as indexed by all the possible values of $\sigma$.  To circumvent the induced difficulties \citep{haghani2020smooth} rounds down the $\sigma_{ul}$ values so that there is a small finite set for each $u$ and hence no explosion in the number of variables or constraints.  

\textbf{Accelerating S-DOI:  }The number of S-DOI grows quadratically (in worst case) in $|\mathcal{N}|$.  To circumvent the enumeration of a quadratic number of $\omega$ variables we may choose to only use some of the S-DOI.  For example we can choose to use only the most restricting S-DOI following the example of \citep{haghani2020smooth} (meaning those $s \in \mathcal{S}$ with the smallest $\rho_{s}$ values).
\subsection{Relaxed Columns}
\label{invalidCOl}
 Pricing is often computationally challenging or even NP-hard\citep{desrosiers2005primer,FlexDOIArticle}.  
 To circumvent solving difficult pricing problems, one approach is to do CG optimization over a super-set of $\Omega$ denoted $\Omega^+$ for which pricing can be solved efficiently over.  Thus we  replace $\Omega$ with $\Omega^+$ in Eq \ref{setcover:cost}, and Eq \ref{sfdoi:cost}. We refer to the additional columns ($\Omega^+\setminus\Omega$) as relaxed.  Considering $\Omega^+$ instead of $\Omega$ may loosen the relaxation.   However if the members of $\Omega^+\setminus\Omega$ are inactive at termination of CG then the solution is equal to that corresponding to optimization over $\Omega$.  
\subsection{Using Invalid DOI}
\label{invalid_doi}
It may be the case that the mechanism that produces DOI is imperfect \citep{Gschwind2016Dual} leading to proposed DOI that cut off all dual optimal solutions.  However the DOI may be intuitive and close to correct.  Thus the proposed DOI, which we refer to as relaxed DOI, while invalid may accelerate optimization of a perhaps slightly weaker LP relaxation.  To ensure that the LP relaxation is not weakened we remove such DOI as needed.  For example in application of SF-DOI we remove from the primal LP any $\omega$ or $\xi$ variables that have non-zero values at termination of CG.  Next we restart optimization using the current set of $\Omega_R$ for initialization of CG.  We repeat this until no DOI are active at termination of CG.  This must terminate since there are a finite number of $\omega$ and $\xi$ terms.

The use of relaxed DOI could make the RMP unbounded in the primal and infeasible in the dual. To correct this primal variables corresponding to DOI should be removed when the RMP would set them to $\infty$. We have not observed unbounded primal RMP objectives/solutions in our experiments. 
\subsection{Application to Vehicle Routing}
\label{vehicleROuting}
We now formulate CVRP as a minimum weight set cover problem.  We use $\{1,2...N\}$ to denote the set of customers and $0,N+1$ to denote the starting and ending depots; where $N$ is the number of customers.  We use $\Omega$ to denote the set of feasible routes, which we index by $l$, each of which starts at the starting depot and ends at the ending depot.  A route is feasible if it contains no customer more than once and services no more demand than it has capacity. We describe $\Omega$ using $a_{ul} \in \{0,1\}$ where $a_{ul}=1$ if and only if route $l$ services customer $u$ and $a_{ul}=0$ otherwise. If a route visits a given customer then that route services the entire demand of that customer.  
We use positive integer $d_{u}$ to denote the number of units of commodity that are demanded by customer $u$. We use the positive integer $K$ to denote the capacity of a single vehicle.  The capacity constraint for a vehicle route is written as follows.
\begin{align}
\label{cap_const_eq}
 \sum_{u \in \mathcal{N}}a_{ul}d_u\leq K \quad \forall l \in \Omega
\end{align}
The cost of any route $l \in \Omega$ is denoted $c_l$ and described below.  We use $T_{uvl}=1$ to indicate that in the route $l$ that customer (or depot) $u$ is followed immediately by customer (or depot) $v$.  We use $c_{uv}$ to denote the associated cost, which is the distance between $u,v$ in metric space.  In CVRP $c_{uv}$ satisfies the triangle inequality.  The cost $c_l$ is a fixed constant $f \in \mathbb{R}_{0+}$ for instantiating the vehicle plus the total distance traveled on route $l$.  The offset $f$ corresponds to a dualized constraint providing an upper bound on the number of vehicles used.  We use $\mathcal{N}=\{1,2,3...N\}$ to denote the set of customers and $\mathcal{N}^+$ to denote the union of the set $\mathcal{N}$, the starting depot $0$ and ending depot $N+1$.  We define $c_l$ formally as:   
    \begin{align}
    c_l=f+\sum_{\substack{u \in \mathcal{N}^+ \\ v \in \mathcal{N}^+ }}T_{uvl}c_{uv} 
    \end{align}

CVRP is commonly attacked as a minimum weight set cover problem using the formulation in  Eq \ref{setcover:cost}.  Finding negative reduce cost columns is attacked as a elementary resource constrained shortest path problem (ERCSPP) \citep{desrosiers2005primer}, which is strongly NP-Hard.  Specifically the computational difficulty of the ERCSPP grows exponentially in $|\mathcal{N}|$. 
\subsection{Relaxed Columns:  ng-routes}
\label{ng_route_overview}
The difficulty of solving the ERCSPP is a consequence of enforcing the elementarity constraint during pricing, which states that no item can be included more than once in a route.  To circumvent the difficulty of enforcing elementarity a common approach is to weaken the  LP relaxation in Eq \ref{setcover:cost} to consider a superset of vehicle routes $\Omega$, which we denote as $\Omega^+$ and refer to as the set of ng-routes \citep{baldacci2011new}.  The ng-route relaxation partially relaxes elementarity by enforced elementarity only between nearby customers as follows.  

Each customer is associated with a subset $\mathcal{N}_u \subset \mathcal{N}$ called its neighborhood, corresponding to its nearest neighbors in metric space.  The size of this neighborhood is a user defined hyper-parameter which trades off tightness of the relaxation and optimization difficulty.  A route lies in the expanded set $\Omega^+$ if the following holds. No cycle within a route starting and ending at $u$ contains exclusively customers for which $u$ is one of their neighbors. Formally let $m_1,m_2$ be positive integers where $1 \leq m_1<m_2 \leq \sum_{u \in \mathcal{N}} a_{ul}$ and let $u_m$ be the $m$'th customer visited in the route $l$.  A route lies in $\Omega^+$ if capacity is not violated (as described by Eq \ref{cap_const_eq}) and $ \forall m_1,m_2$  s.t. $u_{m_1}=u_{m_2}$ there exists a $m_1<m<m_2$ s.t. $u_{m_1} \not \in \mathcal{N}_{u_m}$.  Observe that the presence of cycles within routes means that $a_{ul}$ lies in $\mathcal{Z}_{0+}$ not $\{0,1\}$.

No optimal binary valued solution to Eq \ref{sfdoi:cost}, when optimizing over $\Omega^+$, uses a route in $\Omega^+-\Omega$, since the cost of the solution could be decreased by removing customers from active routes so that no customer is included more than once.
However an optimal fractional solution to  Eq \ref{sfdoi:cost} may include routes in $\Omega^+-\Omega$.  In practice however optimization over $\Omega^+$ does not significantly weaken the relaxation \citep{baldacci2011new}.  Finding the lowest reduced cost ng-route can be efficiently computed via a dynamic program \citep{baldacci2011new}.  

\section{SF-DOI for CVRP}
\label{Sec_DOI_valid_CVRP}
In this section we provide a mechanism to produce valid S-DOI and F-DOI for CVRP where optimization is done over $\Omega$. We use the resultant SF-DOI as relaxed DOI for ng-routes columns.  We provide two types of each of S-DOI and F-DOI one of which is easier to compute, and the other of which is potentially tighter.  We use the later type of S-DOI and F-DOI in experiments. 

We organize this section as follows.  In Section \ref{sec_S_doi_cvrp_elem_easy}, and Section \ref{sec_S_doi_cvrp_elem} we provide the easier to compute and potentially tighter variants of S-DOI respectively. In Section \ref{sec_F_doi_cvrp_elem}, and Section \ref{sec_F_doi_cvrp_elem_2} we provide the easier to compute and potentially tighter variants of F-DOI respectively.  In Section \ref{ng_relation} we apply the SF-DOI in this section as relaxed DOI for ng-routes.  
\subsection{S-DOI:  Easy to Compute}
\label{sec_S_doi_cvrp_elem_easy}
Observe via Eq \ref{cap_const_eq} that $(u,v) \in \mathcal{S}$ if and only if the demand of customer $u$ is greater than or equal to the demand of customer $v$ (meaning $d_u \geq d_v$).  Consider any  $(u,v) \in \mathcal{S}$ and route $l \in \Omega_u-\Omega_v$.  Let $l'$ be the route produced by replacing $u$ with $v$ in $l$ (meaning $l'=$swap$(l,u,v)$).  Let $u_-$,$u_+$ be the customers or depots immediately preceding/succeeding $u$ in $l$.  We write $ c_{l'}-c_l $ below  explicitly. 
\begin{align}
\label{diffEqr}
c_{l'}-c_l = (c_{u_-v}-c_{u_-u})+(c_{vu_+}-c_{uu_+}) 
\end{align} 
Via the triangle inequality $c_{u_-v}\leq c_{u_-u}+c_{uv}$ and $c_{v u_+}\leq c_{uv}+ c_{u u_+}$.  Below we plug in these upper bounds on $c_{u_-v}$ and $c_{v u_+}$ into Eq \ref{diffEqr}.
\begin{align}
  c_{l'}-c_l \leq 2c_{uv}
\end{align}
 Therefor setting $\rho_{uv}=2c_{uv}$ for all pairs of unique elements $u,v$ s.t. $d_u\geq d_v$ satisfies Eq \ref{smoothDOiDef}. 
\subsection{S-DOI:  Tighter Variant}
\label{sec_S_doi_cvrp_elem}
Observe that $\rho_{uv}$ in the context of CVRP is an upper bound on the maximum amount that the cost of a route can increase when replacing customer $u$ with $v$.  As in Section \ref{sec_S_doi_cvrp_elem_easy} $(u,v) \in \mathcal{S}$ if and only if $d_u \geq d_v$.  We write $\rho_{uv}$ below as exactly the maximum amount that the cost of a route can increase when replacing customer $u$ with $v$.  
\begin{align}
\label{s_doi_harder}
\rho_{uv}=    \max_{\substack{u_1 \in \mathcal{N}^+-(N+1)-u-v \\ u_2 \in \mathcal{N}^+-0-u-v-u_1 }} (c_{u_1v}+c_{v u_2})-(c_{u_1u}+c_{uu_2})
\end{align}
Observe that by iterating over all possible pairs $ \in \{0...N+1\},$ we can efficiently evaluate $\rho_{uv}$ in Eq \ref{s_doi_harder}.
\subsection{F-DOI: Easy to Compute}
\label{sec_F_doi_cvrp_elem}
We now consider the constraints needed to satisfy the description of F-DOI in Section \ref{flex_doi_desc}. We set $\sigma_{ul}=0$ when $a_{ul}=0$ and otherwise use the maximum value satisfying Eq \ref{altBoundFDOI}.  Computing the maximum value satisfying Eq \ref{altBoundFDOI} is done exactly by considering all possible predecessor/successor pairs for $u$ constructed from route $l$ and respecting the order of route $l$.  We then connect the predecessor to the successor directly instead of via $u$ in the route created by removing $u$.  We write $\sigma_{ul}$ as optimization below using the following additional notation.  We describe the customers/depots that compose a route $l$ in the order that they are visited in route $l$ from first to last with:    $l=\{u^l_0,u^l_1,u^l_2...u^l_{|\mathcal{N}_l|},u^l_{N+1}\}$.   Note that $u^l_0$,$u^l_{N+1}$ are the starting depot and ending depot respectively.
\begin{align}
\sigma_{ul} & \longleftarrow \min_{\substack{(i, j): i < k < j\\ u=u^l_k}}\left\{c_{u^l_i u} + c_{u u^l_j} - c_{{u^l_{i}}{u^l_{j}}}\right\} & l\in\Omega, 
\end{align}
\subsection{F-DOI:  Tighter Variant}
\label{sec_F_doi_cvrp_elem_2}
We now consider the constraints needed to satisfy the description of F-DOI in Section \ref{flex_doi_desc} using the notation provided in Section \ref{sec_F_doi_cvrp_elem}.    We use $\nu^l_{ij}$ to denote the change in cost incurred by removing all of  $u^l_i,u^l_{i+1},u^l_{i+2}...u^l_{i+j}$ from $l$ and connecting $u^l_{i-1}$ to $u^l_{i+j+1}$ directly. We describe $\nu^l_{ij}$ formally below.   
\begin{align}
    \nu^l_{ij}=c_{u^l_{i-1}u^l_{i+j+1}}-\sum_{n=i}^{i+j+1}c_{u^l_{n-1}u^l_n}
\end{align}
Observe that $\nu$ is non-positive because the triangle inequality holds $c_{uv}$ terms for CVRP. We express $c_{\hat{l}}$ for $\hat{l}=remove(l,\hat{\mathcal{N}})$ for any $\hat{\mathcal{N}} \subseteq \mathcal{N}_l$ using $A^{\hat{l}}_{ij} \in \{0,1\}$, which is defined for all $1\leq i$, $0\leq j$, $i+j\leq |\mathcal{N}_l|$.   Here $A^{\hat{l}}_{ij}=1$ indicates that none of the customers $\{u^l_i,u^l_{i+1},u^l_{i+2}...u^l_{i+j}\}$ are included in $\hat{l}$ but both the customer/depot preceding $u^l_i$ and succeeding $u^l_{i+j}$ are included and otherwise we set $A^{\hat{l}}_{ij}=0$.   Using $\nu,A$ we write $c_{\hat{l}}$ below. 
\begin{align}
\label{defmunewGamma}
    c_{\hat{l}}=c_l+\sum_{\substack{i\geq 1\\ j\geq 0 \\ i+j \leq |\mathcal{N}_l|}}\nu^l_{ij}A^{\hat{l}}_{ij}
\end{align}
We apply Eq \ref{defmunewGamma} to Eq \ref{criter3} then re-order the terms.  
\begin{align}
\label{tmpp1}
    c_l -c_{\hat{l}}\geq \sum_{u \in \hat{\mathcal{N}}}\sigma_{ul}\\
\label{tmpp2}
    c_l \geq c_l+\sum_{\substack{i\geq 1\\ j\geq 0 \\ i+j \leq |\mathcal{N}_l|}}A^{\hat{l}}_{ij}(\nu^l_{ij}+\sum_{n=i}^{i+j}\sigma_{u^l_n l})
\end{align}
It is a necessary and sufficient condition to ensure that Eq \ref{tmpp2} is obeyed that for every $i\geq 1,j\geq0, i+j\leq |\mathcal{N}_l|$ that the following holds.
\begin{align}
\label{suffCOn}
0\geq  \nu^l_{ij}+\sum_{n=i}^{i+j}\sigma_{u^l_n l}
\end{align}
Using Eq \ref{suffCOn} we consider the selection of $\sigma_{ul}$ as optimization.  We seek to maximize $\sum_{u \in \mathcal{N}_l}\sigma_{ul}$ s.t. Eq \ref{suffCOn} is satisfied.  We maximize $\sum_{u \in \mathcal{N}_l }\sigma_{ul}$ in hopes to maximize the ``rebate" we receive when solving the RMP thus decreasing the objective of the current RMP.  We write this below as an LP.
\begin{align}
\label{lpObjGetXi}
    \max_{\sigma \geq 0} \sum_{u \in \mathcal{N}_l}\sigma_{ul}\\
    0\geq  \nu^l_{ij}+\sum_{n=i}^{i+j}\sigma_{u^l_nl} \quad \forall \{ i\geq 1,j\geq0, i+j\leq |\mathcal{N}_l| \} \nonumber 
\end{align}
Solving Eq \ref{lpObjGetXi} is a small linear program with $|\mathcal{N}_l|$ variables and $|\mathcal{N}_l|+{{|\mathcal{N}_l|} \choose 2}$ constraints.  It is desirable that a $\ell_2$ norm be imposed on $\sigma_{ul}$ to produce a solution that does not have extreme valued $\sigma$ variables and hence encourage extreme valued $\alpha$ terms.
To do this we choose to minimize the $\ell_2$ norm of the $\sigma$ terms subject to the constraint that we are near optimal (within a factor $\delta=.999$ of optimality) with respect to Eq \ref{lpObjGetXi}.  We write this as optimization below.  
\begin{align}
\label{qpObjGetXi}
    \min_{\sigma \geq 0} \sum_{u \in \mathcal{N}_l}\sigma_{ul}\sigma_{ul}\\
    0\geq  \nu^l_{ij}+\sum_{n=i}^{i+j}\sigma_{ul} \quad \forall \{ i\geq 1,j\geq0, i+j\leq |\mathcal{N}_l| \} \nonumber\\
     \sum_{u \in \mathcal{N}_l}\sigma_{ul}\geq \delta \mbox{Eq }\ref{lpObjGetXi} \nonumber 
\end{align}

\subsection{Relationship to ng-routes:}
\label{ng_relation}
%
In this section we establish that the DOI described in Section \ref{sec_S_doi_cvrp_elem_easy}-Section \ref{sec_F_doi_cvrp_elem_2} are not valid for ng-routes.  Hence they correspond to relaxed DOI for CG optimization over ng-routes. We establish this by the examples below.

\textbf{F-DOI:  }Consider the ng-route $l=\{0,u,v,u,N+1\}$ where $v \notin \mathcal{N}_u$.  If $v$ is removed from $l$ then the resulting route  $l=\{0,u,u,N+1\}$ clearly is not an ng-route.  Thus the F-DOI described in Section \ref{sec_F_doi_cvrp_elem} and  \ref{sec_F_doi_cvrp_elem_2} are not valid for $\Omega^+$ even though they are for $\Omega$.  

\textbf{S-DOI:  }Again consider the ng-route $l=\{0,u,v,u,N+1\}$ where $v \notin \mathcal{N}_u$.  Let $ d_{v} \geq d_{\hat{v}}$.  Observe that replacing $v$ with $\hat{v}$ produces the route $\{0,u,\hat{v},u,N+1\}$, which is not an ng-route. Thus the S-DOI described in Section \ref{sec_S_doi_cvrp_elem_easy} and  \ref{sec_S_doi_cvrp_elem} are not valid for $\Omega^+$ even though they are for $\Omega$. 

\textbf{Practical Use of SF-DOI with ng-routes: }  While the SF-DOI proposed in  Sections \ref{sec_S_doi_cvrp_elem_easy}-Section \ref{sec_F_doi_cvrp_elem_2} are only valid for $\Omega$ not $\Omega^+$ we argue that they hold approximately for $\Omega^+$ since not all swap/removal operations produce columns that lie outside of $\Omega^+$.  The effectiveness of our SF-DOI for CG optimization over  ng-routes thus becomes an empirical question, which we study in Section \ref{sec_exper}.  As noted in Section \ref{invalid_doi} relaxed DOI are removed as required to ensure that they do not prevent CG from optimally solving optimization over $\Omega^+$.   

We now consider the  computation of $\sigma_{ul}$ for $l \in \Omega^+$ .  We compute $\sigma_{ul}$ terms using Eq \ref{qpObjGetXi} as with any route in $\Omega$, but treat the different copies of any given customer as separate customers. %
This results in different copies of any customer $u$ being associated with separate values $\sigma_{ul}$.
We select the smallest value returned for any given $u$ to define $\sigma_{ul}$ and refer to this selection as the ``smallest value rule".  The use of the smallest value rule ensures that the F-DOI do not trivially induce Eq \ref{sfdoi:cost} to be unbounded when optimization is done over $\Omega^+$ as discussed below.  Consider any $l \in \Omega^+$ and the primal feasible solution to Eq \ref{sfdoi:cost} defined by $\xi_{u\sigma}=M a_{ul}   \quad \forall u \in \mathcal{N},\sigma=\sigma_{ul}$  and $\theta_{l}=M$ where $M= \infty$.  Observe that this solution has primal objective equal  to $-\infty$ if the following holds.
\begin{align}
\label{badEq}
\sum_{u \in \mathcal{N}}a_{ul}\sigma_{ul}> c_l
\end{align}
Observe that for any $l \in \Omega^+$ defining $\sigma_{ul}$ using the smallest value rule applied to any feasible solution to Eq \ref{qpObjGetXi} prevents Eq \ref{badEq} from being satisfied.  
\section{Experiments}
\label{sec_exper}

\begin{figure*}[!hbtp]
	\includegraphics[width=0.45\linewidth]{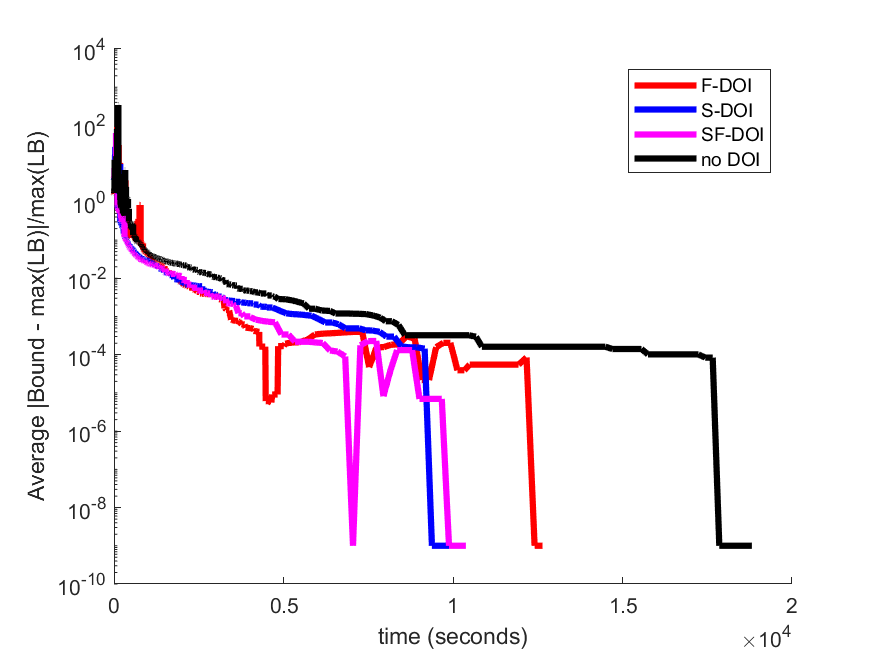}
	\includegraphics[width=0.45\linewidth]{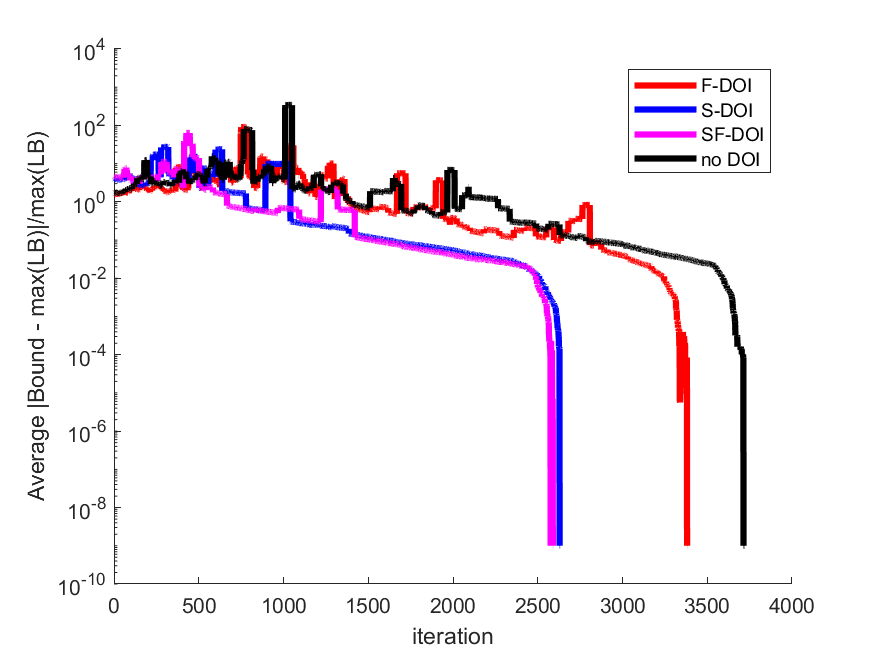}\\
	\caption{Aggregated plots. Relative duality gaps are displayed as relative difference between upper and the maximum lower bounds.
		\textbf{(Left):} Average relative duality gap over all 46 problem instances as a function of time.  
		\textbf{(Right):} Average relative duality gap over all 46 problem instances as a function of iterations.}
	\label{fig:TB1}
\end{figure*}

We test the performance of our DOI on four benchmark CVRP datasets: A, B, P, and E.  Sets A, B, and P were introduced in \citep{augerat1995computational}.  Set E was introduced in \citep{christofides1969algorithm}.  We test on instances with at most 50 customers.  Traversal costs are calculated as the Euclidean distance between customer locations rounded to the nearest integer.  We solve the relaxed ng-routes problem where neighborhoods are set as the five nearest customers.  Pricing amounts to solving an ng-route  shortest path problem, which we solve as a dynamic program.

We evaluate DOI by the speedup in convergence they provide in comparison to non-stabilized CG.  Algorithms are coded in MATLAB and CPLEX is used as our LP-solver.  Experiments are run on an 8-core AMD Ryzen 1700 CPU @3.0 GHz with 32 GB of memory running Windows 10.

Computational results on all 46 problem instances are detailed in Table \ref{table:TB1}.  Aggregate plots showing the average normalized duality gap over all instances as a function of iteration and time are shown in Figure \ref{fig:TB1}.  We see that the S-DOI and SF-DOI offer an average speedup of 20\%.  The S-DOI provide a positive speedup in 44 out of 46 instances, while the SF-DOI (using both F-DOI and S-DOI) provide a positive speedup in 41 out of 46 instances.  The F-DOI did not produce any average speedup over the instances.  Most of the speedup of SF-DOI can be attributed to the S-DOI, however the SF-DOI outperform the S-DOI in 21 out of 46 instances.

The process of removing active DOI at termination as described in Section \ref{invalid_doi} is observed to be a necessary component for convergence.  The S-DOI required removal of active DOI in 2 out 46 instances, while the F-DOI and SF-DOI both required removal of active DOI in 42 out of 46 instances.  The phenomena of DOI inducing unbounded RMP primal solutions is not observed in our experiments.

\begin{table*}[t!]
	\centering
	\scalebox{0.8}{
		\begin{tabular}{|c|c c c c|c c c|} 
		\hline
		& \multicolumn{4}{c|}{\bf Time (sec)} & \multicolumn{3}{c|}{\bf Speedup}\\
		\bf Instance & \bf Standard & \bf S-DOI & \bf F-DOI & \bf SF-DOI & \bf S-DOI & \bf F-DOI & \bf SF-DOI \\
		\hline
    		A-n32-k5 & 387 & 339 & 495 & 333 & 1.1 & 0.8 & 1.2 \\
    		A-n33-k5 & 339 & 258 & 301 & 250 & 1.3 & 1.1 & 1.4 \\
    		A-n33-k6 & 371 & 230 & 299 & 231 & 1.6 & 1.2 & 1.6 \\
    		A-n34-k5 & 401 & 324 & 415 & 376 & 1.2 & 1.0 & 1.1 \\
    		A-n36-k5 & 523 & 454 & 490 & 516 & 1.2 & 1.1 & 1.0 \\
    		A-n37-k5 & 804 & 597 & 832 & 734 & 1.3 & 1.0 & 1.1 \\
    		A-n37-k6 & 450 & 433 & 476 & 375 & 1.0 & 0.9 & 1.2 \\
    		A-n38-k5 & 734 & 499 & 651 & 539 & 1.5 & 1.1 & 1.4 \\
    		A-n39-k5 & 731 & 631 & 631 & 604 & 1.2 & 1.2 & 1.2 \\
    		A-n39-k6 & 695 & 522 & 552 & 569 & 1.3 & 1.3 & 1.2 \\
    		A-n44-k5 & 931 & 748 & 783 & 761 & 1.2 & 1.2 & 1.2 \\
    		A-n45-k5 & 1138 & 947 & 1008 & 979 & 1.2 & 1.1 & 1.2 \\
    		A-n45-k5 & 938 & 778 & 894 & 807 & 1.2 & 1.0 & 1.2 \\
    		A-n46-k6 & 1128 & 865 & 1153 & 976 & 1.3 & 1.0 & 1.2 \\
    		A-n48-k5 & 1260 & 970 & 1213 & 987 & 1.3 & 1.0 & 1.3 \\
    		B-n31-k5 & 386 & 237 & 410 & 355 & 1.6 & 0.9 & 1.1 \\
    		B-n34-k5 & 552 & 462 & 503 & 395 & 1.2 & 1.1 & 1.4 \\
    		B-n35-k5 & 585 & 494 & 545 & 420 & 1.2 & 1.1 & 1.4 \\
    		B-n38-k6 & 741 & 623 & 682 & 484 & 1.2 & 1.1 & 1.5 \\
    		B-n39-k5 & 951 & 831 & 762 & 797 & 1.1 & 1.2 & 1.2 \\
    		B-n41-k6 & 804 & 534 & 774 & 532 & 1.5 & 1.0 & 1.5 \\
    		B-n43-k6 & 1207 & 923 & 1149 & 920 & 1.3 & 1.1 & 1.3 \\
    		B-n44-k7 & 945 & 704 & 850 & 937 & 1.3 & 1.1 & 1.0 \\
    		B-n45-k5 & 2064 & 1750 & 1999 & 1409 & 1.2 & 1.0 & 1.5 \\
    		B-n45-k6 & 1377 & 964 & 1235 & 1002 & 1.4 & 1.1 & 1.4 \\
    		B-n50-k7 & 2151 & 1383 & 1600 & 1377 & 1.6 & 1.3 & 1.6 \\
    		B-n50-k8 & 1386 & 1356 & 1264 & 1272 & 1.0 & 1.1 & 1.1 \\
    		B-n51-k7 & 1836 & 1597 & 1969 & 1494 & 1.1 & 0.9 & 1.2 \\
    		E-n22-k4 & 66 & 49 & 78 & 55 & 1.4 & 0.8 & 1.2 \\
    		E-n23-k3 & 18815 & 10032 & 12641 & 10380 & 1.9 & 1.5 & 1.8 \\
    		E-n30-k3 & 996 & 760 & 964 & 1058 & 1.3 & 1.0 & 0.9 \\
    		E-n33-k4 & 4732 & 4677 & 5006 & 4024 & 1.0 & 0.9 & 1.2 \\
    		E-n51-k5 & 4014 & 3372 & 3324 & 3510 & 1.2 & 1.2 & 1.1 \\
    		P-n16-k8 & 0 & 0 & 2 & 2 & 0.9 & 0.1 & 0.2 \\
    		P-n19-k2 & 133 & 113 & 118 & 119 & 1.2 & 1.1 & 1.1 \\
    		P-n20-k2 & 201 & 169 & 166 & 164 & 1.2 & 1.2 & 1.2 \\
    		P-n21-k2 & 245 & 221 & 245 & 178 & 1.1 & 1.0 & 1.4 \\
    		P-n22-k2 & 309 & 298 & 249 & 227 & 1.0 & 1.2 & 1.4 \\
    		P-n22-k8 & 8 & 7 & 13 & 14 & 1.2 & 0.6 & 0.6 \\
    		P-n23-k8 & 4 & 4 & 8 & 7 & 1.2 & 0.5 & 0.6 \\
    		P-n40-k5 & 1116 & 972 & 930 & 926 & 1.1 & 1.2 & 1.2 \\
    		P-n45-k5 & 1999 & 1813 & 2005 & 1438 & 1.1 & 1.0 & 1.4 \\
    		P-n50-k7 & 569 & 442 & 550 & 494 & 1.3 & 1.0 & 1.2 \\
    		P-n50-k8 & 1779 & 1442 & 1603 & 1532 & 1.2 & 1.1 & 1.2 \\
    		P-n50-k10 & 839 & 875 & 1015 & 951 & 1.0 & 0.8 & 0.9 \\
    		P-n51-k10 & 650 & 573 & 590 & 451 & 1.1 & 1.1 & 1.4 \\
		\hline
		\bf mean & \bf1354  & \bf1006  & \bf1162 & \bf999 & \bf1.2 & \bf1.0 & \bf1.2 \\
		\bf median & \bf773 & \bf610 & \bf721 & \bf587 & \bf1.2 & \bf1.1 & \bf1.2 \\
		\hline
	\end{tabular}}
	\caption{CVRP runtime results}
	\label{table:TB1}
\end{table*}
\section{Conclusions}
\label{sec_conc}
In this document we adapt smooth and flexible dual optimal inequalities (SF-DOI) \citep{haghani2020smooth} to accelerate the convergence of column generation when applied to minimum weight set covering based formulations with relaxed columns. We apply our approach to the capacitated vehicle routing problem, which we formulate as a minimum weight set cover problem over ng-routes. We demonstrate significant improvements in the speed of optimization with no weakening of the underlying relaxation.  
In future work we seek to extend our approach to operate in the context of branch and price \citep{barnprice}.  We also seek to extend our approach to consider the application of valid inequalities such as subset-row inequalities \citep{jepsen2008subset}, which are used to tighten the set cover linear programming relaxation.  Another key avenue of future research is to apply SF-DOI to vehicle routing problems with time windows.
%

\bibliography{col_gen_bib}
\bibliographystyle{icml2020}
\appendix
\section{Proof of Sufficient F-DOI Criteria: }
\label{appendProof}
We  now prove that Eq \ref{altBoundFDOI} is a sufficient condition to satisfy  Eq \ref{criter3}.  Let $\hat{l}_{u},\hat{l}_{-u},\sigma_{ul}$ be defined as as the arg mimizer and minimizer of the right hand side (RHS) of Eq \ref{altBoundFDOI} as follows.
\begin{align}
\label{defhatlu}
    \hat{l}_{u} = \mbox{arg} \min_{\substack{\hat{l} \in \Omega_u \\ N_{\hat{l}}\subseteq \mathcal{N}_l }}c_{\hat{l}}-c_{\hat{l}_{-u}}\quad \quad  \mbox{where} \quad \hat{l}_{-u}=\mbox{remove}(\hat{l},\{u\})\\
    \sigma_{ul}=c_{\hat{l}_u}-c_{\hat{l}_{-u}} \nonumber 
\end{align}
  Let us remove items in $\hat{\mathcal{N}}$ from $l$ in an arbitrary order.  Let $u_k$ be the $k$'th member in $ \hat{\mathcal{N}}$. Let $l_k$ refer to the column constructed by removing the first $k$ items in $\hat{\mathcal{N}}$ from $l$. Observe that $l_0=l$ and $l_{|\hat{\mathcal{N}}|}=l'$.  We now rewrite  Eq \ref{criter3} using Eq \ref{defhatlu} to define $\sigma_{ul}$ and also add to the RHS $0=\sum_{k=1}^{|\hat{N}|-1}c_{l_k}-c_{l_k}$. 
\begin{align}
\label{eq_check}
    \sum_{u\in \hat{\mathcal{N}}}\sigma_{ul}=\sum_{u\in \hat{\mathcal{N}}}c_{\hat{l}_u}-c_{\hat{l}_{-u}}\leq c_l - c_{l'}=\sum_{k=0}^{|\mathcal{N}_l|-1}c_{l_k}-c_{l_{k+1}}
\end{align}
 Observe that  $c_{\hat{l}_{u_k}}-c_{\hat{l}_{-u_k}} \leq c_{l_{k-1}}-c_{l_{k}}$ by definition in Eq \ref{defhatlu} for all $1\leq k \leq |\hat{\mathcal{N}}|$.  Thus Eq \ref{eq_check} is valid by construction and therefor Eq \ref{criter3} is satisfied by  Eq \ref{altBoundFDOI}.
\end{document}